\documentclass[letterpaper, 10 pt, conference]{ieeeconf}  

\IEEEoverridecommandlockouts                              

\usepackage{cite}
\usepackage{amsmath,amssymb,amsfonts}
\usepackage{algorithmic}
\usepackage{graphicx}
\usepackage{textcomp}
\usepackage{xcolor}
\usepackage{fancyhdr}
\fancypagestyle{withfooter}{
  
  \fancyfoot[C]{\footnotesize Accepted as Extended Abstract to the IEEE ICRA@40 2024}
}

\title{\LARGE \bf
Learning to Turn: Diffusion Imitation for Robust Row Turning in Under-Canopy Robots}

\author{Arun N. Sivakumar$^{1}$, Pranay Thangeda$^{2}$, Yixiao Fang$^{1}$, Mateus V. Gasparino$^{1}$,\\ Jose Cuaran$^{1}$, Melkior Ornik$^{2}$, Girish Chowdhary$^{1}$
\thanks{$^{1}$ Field Robotics Engineering and Sciences Hub (FRESH), University of Illinois Urbana-Champaign}%
\thanks{$^{2}$Department of Aerospace Engineering, University of Illinois Urbana-Champaign}%
\thanks{This work was supported in part by NSF STTR \#1951250, NSF NRI 2.0 NIFA \#2021-67021-33449, AIFARMS \#1024178, NSF-USDA COALESCE \#2021-67021-34418, USDA grants iCOVER(\#NR233A750004G066) and iFARM(\#2022-77038-37306).}
}

\begin{document}

\maketitle
\thispagestyle{withfooter}
\pagestyle{withfooter}

\begin{abstract}
Under-canopy agricultural robots require robust navigation capabilities to enable full autonomy but struggle with tight row turning between crop rows due to degraded GPS reception, visual aliasing, occlusion, and complex vehicle dynamics. We propose an imitation learning approach using diffusion policies to learn row turning behaviors from demonstrations provided by human operators or privileged controllers. Simulation experiments in a corn field environment show potential in learning this task with only visual observations and velocity states. However, challenges remain in maintaining control within rows and handling varied initial conditions, highlighting areas for future improvement.

\end{abstract}



\section{INTRODUCTION}
Under-canopy agricultural robots have the potential to autonomously navigate the narrow gap between crop rows for plant-level monitoring and care. However, their development has been hindered by challenges in reliable localization and navigation within these occluded, visually homogeneous environments \cite{winterhalter2021localization}. While some progress has been made on crop row following using classical computer vision \cite{winterhalter2021localization,aastrand2005vision} and learning \cite{sivakumar2021learned} techniques, executing precise turns to transition between rows remains an open challenge \cite{gasparino2023cropnav}. 

In open fields, agricultural robots typically rely on GPS for localization and navigation \cite{gasparino2023cropnav}. However, GPS signal quality severely degrades under dense crop canopies, necessitating alternative navigation methods \cite{higuti2019under}. While depth sensors can help, they are expensive, sensitive to clutter, and increase system complexity \cite{gao2020spraying}. Monocular RGB vision is appealing for its simplicity and low cost, but turning is still difficult due to the need to execute a tight maneuver in a highly occluded scene with strong visual aliasing between the rows. The complex dynamics of under-canopy robots with limited clearance further complicate the control problem.

In this work, we propose to learn row turning policies from demonstrations using imitation learning, specifically leveraging the recent work diffusion models for policy synthesis \cite{janner2022planning}, \cite{chi2023diffusionpolicy}. Imitation learning is well-suited for acquiring complex skills that are difficult to specify with hand-engineered controllers \cite{pomerleau1988alvinn} and can utilize demonstrations from human teleoperators or privileged automated controllers. By including recovery behaviors in the demonstrations, the learned policy can better handle a wider range of states, including those that may occur due to gradual drift or minor errors during operation. This expanded state coverage helps the policy remain robust and effective even when faced with situations that deviate from ideal conditions.

Diffusion models learn a denoising process from data that can flexibly model complex distributions. While originally developed for image generation, recent work has shown how to leverage them as policy representations for robot learning \cite{janner2022planning}, \cite{chi2023diffusionpolicy}. We employ a conditional diffusion model to represent the row turning policy, where the conditioning input is the RGB image and the output is the action distribution. 

We evaluate our approach using a high-fidelity simulator that models under-canopy navigation dynamics. This enables extensive testing of our learned policies across diverse scenarios prior to real-world deployment.

\section{APPROACH}
Let $\mathcal{O}$ be the space of RGB camera observations from one or more onboard cameras and $\mathcal{A} \subseteq \mathbb{R}^m$ the space of robot actions. The goal is to learn a policy $\pi: \mathcal{O} \to \mathcal{P}(\mathcal{A})$ mapping observations to a distribution over actions to execute the row turn. We collect a dataset $\mathcal{D} = \{(\mathbf{o}_i, \mathbf{a}_i)\}_{i=1}^{N}$ of observation-action trajectories from demonstrations.

\begin{figure*}[t]
    \centering
    \includegraphics[width=0.85\textwidth]{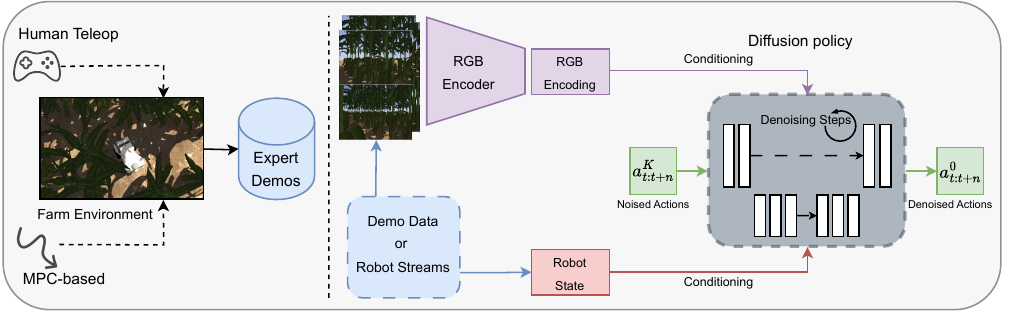}
    \caption{Overview of the proposed method for learning row turning behaviors using diffusion policies. (Left) Demonstrations collected in the simulation environment using human teleoperation and procedurally generated demonstrations that utilize privileged information. (Right) Architecture that takes in RGB and robot state observation history and generates sequence of actions for execution. }
    \label{fig:method_overview}
\end{figure*}

We employ a conditional denoising diffusion probabilistic model (DDPM) \cite{ho2020denoising,janner2022planning} to represent $\pi$. The DDPM comprises a forward noising process that gradually adds Gaussian noise to an action sequence and a learned reverse process that attempts to denoise the sequence when conditioned on an observation. Formally, the forward process is:
\begin{equation}
q(\mathbf{a}_t | \mathbf{a}_{t-1}, \mathbf{o}) = \mathcal{N}(\mathbf{a}_t; \sqrt{1 - \beta_t} \mathbf{a}_{t-1}, \beta_t \mathbf{I})
\end{equation}
where $t \in \{1, \dots, T\}$ indexes the timestep, $\beta_t \in (0, 1)$ is a variance schedule, and $\mathbf{I}$ is the identity matrix. The reverse process $p_\theta$ is parameterized by a neural network with parameters $\theta$ that predicts the noise added at each step:
\begin{equation}
p_\theta(\mathbf{a}_{t-1} | \mathbf{a}_t, \mathbf{o}) = \mathcal{N}(\mathbf{a}_{t-1}; \mathbf{\mu}_\theta(\mathbf{a}_t, \mathbf{o}, t), \mathbf{\Sigma}_\theta(\mathbf{a}_t, \mathbf{o}, t))
\end{equation}
where $\mathbf{\mu}_\theta$ and $\mathbf{\Sigma}_\theta$ are the mean and covariance predicted by the network. The network is trained to maximize the variational lower bound on the log likelihood $\log p_\theta(\mathbf{a} | \mathbf{o})$. The training objective is:
\begin{align}
\mathcal{L}(\theta) &= \mathbb{E}_{q(\mathbf{a}_{0:T} | \mathbf{o})} \Bigl[ \log p_\theta(\mathbf{a}_0 | \mathbf{a}_1, \mathbf{o}) \nonumber \\ 
&- \sum_{t=2}^{T} \mathrm{D_{KL}}[q(\mathbf{a}_{t-1}|\mathbf{a}_t, \mathbf{o}) \| p_\theta(\mathbf{a}_{t-1}|\mathbf{a}_t, \mathbf{o})] \Bigr]
\end{align}

To sample from the policy, we first sample $\mathbf{a}_T \sim \mathcal{N}(\mathbf{0}, \mathbf{I})$ and then iteratively sample $\mathbf{a}_{t-1} \sim p_\theta(\cdot | \mathbf{a}_t, \mathbf{o})$ for $t = T, \dots, 1$ using the learned reverse process.

\section{EXPERIMENTS}
We train and evaluate our approach in a simulated cornfield environment using Gazebo. Our platform is Terrasentia \cite{zhang2020high}, a wheeled under-canopy agricultural robot that uses a skid-steer drive mechanism. The robot is equipped with three monocular RGB cameras: one on the front and one on each side. We control the robot using a two-dimensional action space consisting of linear and angular velocity.
\begin{figure}[h]
    \centering
    \includegraphics[width=0.95\columnwidth]{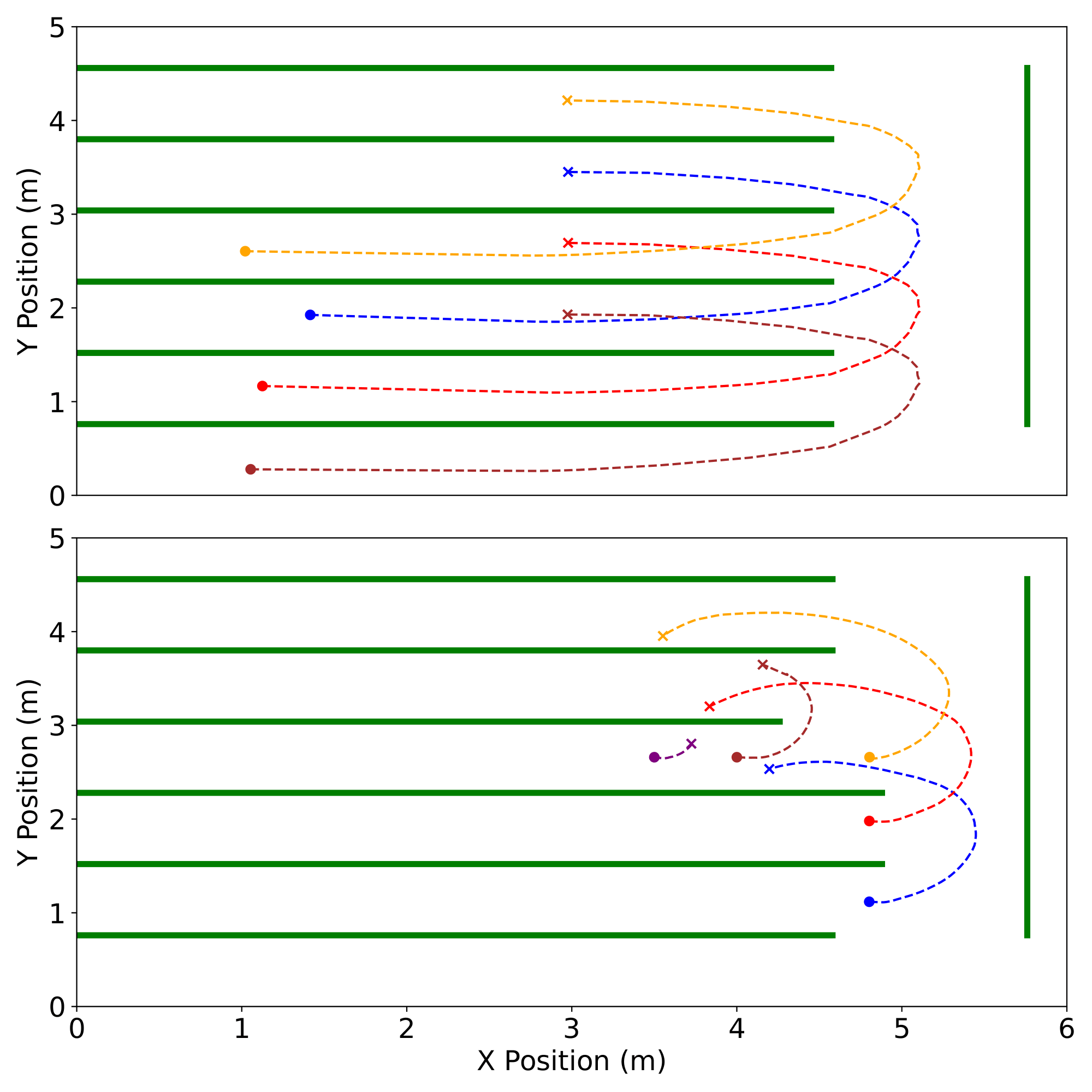}
    \caption{Bird's eye view visualization of trajectories in a corn field. Rows of corn are represented using green solid lines. Each trajectory is represented by a unique color, with \textbullet  indicating the starting point and $\times$ marking the end point. (Top) Sample expert demonstration trajectories collected for training the policy. Multiple trajectories showcase the diversity of paths used in the training data. (Bottom) Sample rollouts generated by the trained policy for various initial conditions, illustrating the policy's ability to produce trajectories similar to the demonstration data.}
    \label{fig:results}
\end{figure}
We collect demonstrations from two sources: a human operator using a joystick controller and a model predictive controller (MPC) with access to privileged state information. Our final objective is to transfer the proposed system to the real world, where only human demonstrations are feasible as a source of supervision. This motivation drives our decision to experiment with expert data from both MPC and human operators. We randomize the environment by using different corn models in the simulator. Our dataset comprises 350 demonstration trajectories. Importantly, the human operator demonstrations include scenarios where the robot crashes into corn and recovers from failures. We hypothesize that these scenarios can improve the trained policy's robustness to domain shift during deployment. In this work, we focus on the specific task of turning toward the left direction and skipping one crop row.

Fig. \ref{fig:results} illustrates comparison trajectories between demonstration trajectories from the MPC and evaluation rollouts from the learned policy, which we execute in a closed-loop manner during testing. When initialized at the end of the row, our learned policy successfully tracks the desired trajectory to enter the desired row but fails after entering the row. When initialized before the end of the row, the policy fails by applying excessive angular velocities towards the left. This brittle behavior within the corn rows might stem from minimal variance in control output for these scenarios. The policy appears to reach a suboptimal minimum by primarily learning to minimize control prediction error during turns, where angular velocity values are high. Consequently, it fails to adequately learn appropriate behaviors for states before or after the turn.



\section{FUTURE WORK}
Our simulation experiments demonstrate the potential of imitation learning with diffusion models for autonomous row turning in under-canopy agricultural robots. While showing promise in executing turn maneuvers, further research is needed to fully realize this approach's capabilities. Future work will focus on goal conditioning to flexibly specify different turning directions and target rows. We also plan to deploy the proposed approach on real under-canopy robots and integrate it with row following for end-to-end navigation. 

\bibliographystyle{IEEEtran}
\bibliography{references}

\end{document}